\pdfoutput=1

\documentclass[11pt]{article}

\usepackage[]{acl}

\usepackage{times}
\usepackage{latexsym}
\usepackage{amssymb}
\usepackage{booktabs}
\usepackage{amsmath}
\usepackage{xspace}
\usepackage{graphicx}
\usepackage{algorithm}
\usepackage{algpseudocode}
\usepackage{caption}
\usepackage{subcaption}
\usepackage{stmaryrd}
\usepackage[subtle]{savetrees}

\newcommand{\Vp}{\mathcal{V}_\prompt}
\newcommand{\gen}{\texttt{Gen}\xspace}
\newcommand{\prompt}{m\xspace}

\usepackage[T1]{fontenc}

\usepackage[utf8]{inputenc}

\usepackage{microtype}

\title{Reverse-Engineering Decoding Strategies Given\\ Blackbox Access to a Language Generation System}

\author{Daphne Ippolito\thanks{\hspace{.5em}Google Deepmind, \textdagger University of Toronto}\\
  {\tt dei@google.com} \\\And
  Nicholas Carlini\textsuperscript{*}\\
  {\tt ncarlini@google.com} \\\And
  Katherine Lee\textsuperscript{*}\\
  {\tt katherinelee@google.com} \\
  \AND
  Milad Nasr\textsuperscript{*}\\
  {\tt miladnasr@google.com} \\\And
  Yun William Yu\textdagger\\
  {\tt ywyu@math.toronto.edu} \\}
\date{}

\begin{document}
\maketitle
\begin{abstract}
Neural language models are increasingly deployed into APIs and websites that allow a user to pass in a prompt and receive generated text.
Many of these systems do not reveal generation parameters.
In this paper, we present methods to reverse-engineer the decoding method used to generate text (i.e., top-$k$ or nucleus sampling).
Our ability to discover which decoding strategy was used has implications for detecting generated text. Additionally, the process of discovering the decoding strategy can reveal biases caused by selecting decoding settings which severely truncate a model's predicted distributions.
We perform our attack on several families of open-source language models, as well as on production systems (e.g., ChatGPT).
\end{abstract}

\section{Introduction}
Language models are increasingly being incorporated into web applications and other user-facing tools.%
\footnote{E.g., see \url{https://gpt3demo.com/} for a list of such apps.}
These applications typically do not provide direct access to the underlying language model or the decoding configuration used for generation. 
In this paper, we show how even in this blackbox setting, it is possible to identify the decoding strategy employed for generation.
We consider the case where one only has access to a system that inputs a prompt and outputs a generated response.
We present algorithms for distinguishing the two most popular decoding strategies, top-$k$ and nucleus sampling (a.k.a. top-$p$), and estimating their respective hyperparameters ($k$ and $p$).

The choice of decoding strategy---the algorithm used to sample text from a language model---has a profound impact on the randomness of generated text, introducing biases toward some word choices.
For example, when OpenAI's ChatGPT,\footnote{\url{https://openai.com/blog/chatgpt/}} a chatbot built with large language models, is repeatedly passed a prompt asking it to report the outcome of rolling a twenty-sided die, we found that it only returns 14 of the 20 options, even though all should be equally likely.

Prior work has shown that knowing the decoding method makes it easier to detect whether a writing sample was generated by a language model or else was human-written
\citep{ippolito2020automatic}.
As generated text proliferates on the web, in student homework, and elsewhere, this disambiguation is becoming increasingly important.

Concurrent work to ours by \citet{naseh2023risks} has developed similar strategies for detecting decoding strategy from a blackbox API: however, they focus more on identifying hybrid decoding strategies (including beam search), whereas we focus more on prompt engineering to produce close-to-uniform token distributions that reduce the number of queries needed.
Our proposed methods complement but are not comparable to those of \citet{tay2020reverse}.
Their method trains classifiers that input a generated text sequence and output a prediction for the decoding strategy used to generate it.
In contrast, our method interacts with an API and does not require any data or ML training.

\section{Background}
Neural language models are not inherently generative.
A causal language model $f_\theta$ takes as input a sequence of tokens $x_1, \ldots, x_{t-1}$ and outputs a score for each possible next token $x_t$, computing the a likelihood score for each token in the vocabulary, which can be transformed into a probability distribution by applying a softmax such that
$\text{Prob}(x_t | x_1, \ldots, x_{t-1}) \sim f_\theta (x_1, \ldots, x_{t-1})$.

A \textbf{decoding method} takes this probability distribution as input and samples a particular token to output.
The simplest algorithm is \emph{argmax decoding} (also known as `greedy decoding'), where the most likely next token is outputted. 
Argmax is rarely used in practice because (1) only one generation can be produced for any given prompt, and (2) generations with argmax tend to be repetitive and low-quality.
%

Most commonly used decoding algorithms are based on random sampling:
a token is chosen with probability proportional to the likelihood assigned to it by the model.
Whereas argmax sampling has too little randomness, purely random sampling over the full distribution can have too much, leading to text that is too erratic and prone to errors.
Thus, it is common to modify the distribution to reduce entropy before sampling from it.

In this short paper, we focus on two popular strategies researchers have developed for decoding: top-$k$ sampling \citep{fan2018hierarchical} and top-$p$ sampling \citep{holtzman2019curious} (also known as nucleus sampling).
Top-$k$ sampling involves the implementer picking a fixed hyperparemter $k$ then only ever sampling from the $k$ most likely items by assigning all other items a score of 0 before applying the softmax.
Top-$p$ sampling involves the implementer picking a fixed hyperparamter $p$.
Then at each step $t$ of generation, a $k_t$ is selected such that the $k_t$ most likely vocabulary items cover $p$ proportion of the total probability mass in the distribution.
More precisely, let the notation $x^{(l)}$ refer to the $l$th most likely token in the distribution predicted at step $t$.
We set $k_t$ to the first value for which
$\sum_{l=1}^{k_t} \text{Prob}(x_t=x^{l} | x_1, \ldots, x_{t-1}) \geq p$.
Then, the distribution is truncated to the $k_t$ most likely tokens, as described above for top-$k$.

Other common methods like beam search and temperature annealing are omitted in the interest of space (cf. \citet{zarriess2021decoding} and \citet{wiher2022decoding}).
Temperature annealing simply modifies the probability distributions of the output tokens, so the methods in this manuscript can be easily generalized (and indeed were in the concurrent work of \citet{naseh2023risks}).
Beam search is a bit more complicated, as tokens are not chosen independently of previous tokens; instead, multiple candidate token paths are retained.
As such, it would be necessary to generate more than a single word for each prompt, which is the primary interrogative tool we use here.

\section{Method}
\subsection{Threat Model}
We assume black-box, query-only access to the system $\gen : m \mapsto r$ which takes as input a prompt string $m$ and outputs a textual response $r$; without loss of generality, we assume that the response $r$ is exactly one token long.
The adversary can input arbitrary prompts and observe the output response. 
In most of our experiments, we assume $\gen$ passes $m$ into the language model without any modification, then generates a continuation using an unknown decoding strategy.
However, in some cases, such as for ChatGPT, the system might modify the provided prompt, $m$, such as by prepending few-shot examples, before passing it to the language model. Still, we assume that the causal language model can be repeatedly queried by a fixed prompt $m'$, even if modified from the original $m$.

The adversary's attack objective is to determine the decoding strategy employed by $\gen$, outputting either \texttt{topk} or \texttt{topp}, as well as the value for either $p$ or $k$.

\subsection{Intuition for Method}

We begin with the intuition of our attack.
Suppose we were given a prompt $m$, such that the output of $\gen(m)$ is equally likely to be any item from a set of vocabulary items $\Vp\subseteq\mathcal{V}$.
For example, the prompt ``\texttt{List of capital English letters, chosen uniformly at random:}'' ought to result in the model emitting each of the 26 letters of the alphabet with equal probability.
However, suppose that when we repeatedly prompt the model in this way, it only ever emits 10 different letters.
What could cause this?

One explanation could be that our prompt does not actually induce a uniform probability distribution over each of the 26 letters,
and in fact that the model assigns (nearly) zero probability mass to the $11$th most likely token.
Suppose we know for a fact the prompt does induce a near-uniform distribution on all publicly-available language models:
then the more likely explanation would be that the sampling algorithm itself truncated this distribution---either with top-$k$ or top-$p$ sampling.
By measuring what fraction of the words we would expect to get generated actually do get generated for prompts with known output distributions, we can estimate values for $k$ and $p$ and distinguish between these two techniques.

\subsection{Estimating $k$}
\label{section:estimate_k}
Suppose, for a given prompt $\prompt$, we call $\gen(\prompt)$, $n$ number of times, each time keeping just the first token of the output.
We can trivially lower bound $k$ by observing the number of unique items in a set of responses.
As $n$ approaches $\infty$, all $k$ allowed responses will be observed.
To achieve a compute-efficient attack, our goal is to estimate $k$ while keeping $n$ as small as possible.
Appendix \ref{app:topk} gives theoretical accuracy/runtime estimates for this approach by posing it as the coupon collector problem \cite{polya1930wahrscheinlichkeitsaufgabe}.

In practice, we use Algorithm \ref{alg:topk} (see Appendix), which repeatedly estimates a lower bound for $k$ using two different prompts $\prompt_1$ and $\prompt_2$ for increasing numbers of trials until (1) the two estimates match and (2) the $x^{(k)}$ token appears at least twice in both generations (to prevent spurious matching).

\subsection{Estimating $p$}
\label{section:estimate_p}
In this paper, we set a goal of determining $p$ to within $0.05$ of the true value.
We can upper bound $p$ by constructing a prompt that yields a known, computable distribution over a set of vocab items $\Vp$.
Then to attack a system, we repeatedly sample with the prompt, and count how many of those items are generated.
Let's call this value $k$.\footnote{This is a slight abuse of notation since we used $k$ earlier for top-$k$, but in both cases, this value corresponds precisely to the number of unique tokens seen.}
We estimate $p$ as the sum of the probabilities of the $k$ most likely tokens in the known distribution over items in $\Vp$.
Because our guessed distributions are not perfect, instead of relying on just one prompt for our estimate, we instead average over two upper bounds of $p$ derived from two different prompts.
Although our experiments here use only two prompts, increased precision is achievable by using additional prompts.
The detailed algorithm can be found in the Appendix.

\subsection{Distinguishing Top-$k$ from Top-$p$}

To distinguish between top-$k$ and top-$p$, we need only reject the hypothesis that top-$k$ is used.
It turns out that we can simply reuse Algorithm \ref{alg:topk} because we already built in a measure of concordance in the $k$ predictions.
If the two prompts used as input to Algorithm \ref{alg:topk} continue to yield very different predictions of $k$ no matter how many samples are taken, we can reject the hypothesis of top-$k$ being used.
For rejecting top-$k$, we found it useful to start with two prompts with radically different distributions; it suffices to choose prompts that with very differently sized $\Vp$, such as \textsc{Adverbs} and \textsc{Months}.

Although we did not explore it in this short paper, we could in theory also reject top-$p$ by looking at how closely the $p$ estimates from different prompts match.
This may prove helpful if we wish to determine that neither top-$p$ or top-$k$ is being used, but is unnecessary for simply disambiguating the two.

\begin{table}[]
    \centering
    \small
    \caption{Prompts used for top-$k$ and (top) top-$p$ (bottom) estimation on open-source models. The first two prompts include randomly selected exemplars (shown in blue). For \textsc{Months}, Ramadan is included as the 13th month.}
    \label{tab:prompts}
    \begin{tabular}{p{0.42in}|p{1.8in}|p{0.2in}}
        \toprule
        Name & Prompt & |$\Vp$| \\
        \midrule
        \textsc{Nouns} & List of nouns chosen completely randomly: \textcolor{blue}{dog, slash, altar} & 8,432 \\
        \textsc{Adverbs} & List of adverbs chosen completely randomly: \textcolor{blue}{formally, blatantly, sadly} & 504 \\
        \midrule
        \textsc{Months} & She came to visit in the month of & 13 \\
        \textsc{Dates} & The accident occurred on March & 31 \\
        \bottomrule
    \end{tabular}
\end{table}

\paragraph{Prompt Selection}
In addition to the distributional properties described above, we also need our prompts to have the property that the first space-separated word in the output of $\gen(\prompt)$ is in-fact a word in the vocabulary.
Since we often do not know which vocabulary was used by the model we are attacking,
we choose prompts which yield distributions over words which are likely to be tokens in all models trained on webtext\footnote{Other prompts may be needed for attacking code models.}.
Table \ref{tab:prompts} shows all the prompts used in all experiments except for those we used on ChatGPT (which had to be longer), and Appendix \ref{app:prompt_engineering} gives more details on prompt selection (including for ChatGPT).

\section{Experiments}
We conduct experiments on four language models where we can set the decoding strategy: GPT-2 Base and XL \citep{radford2019language}, GPT-3 Davinci \citep{brown2020language}, BLOOM 3B \citep{scao2022bloom}, and Pythia 2.7B \citep{2023pythia}.

\begin{figure}
    \centering
    \includegraphics[width=0.8\linewidth]{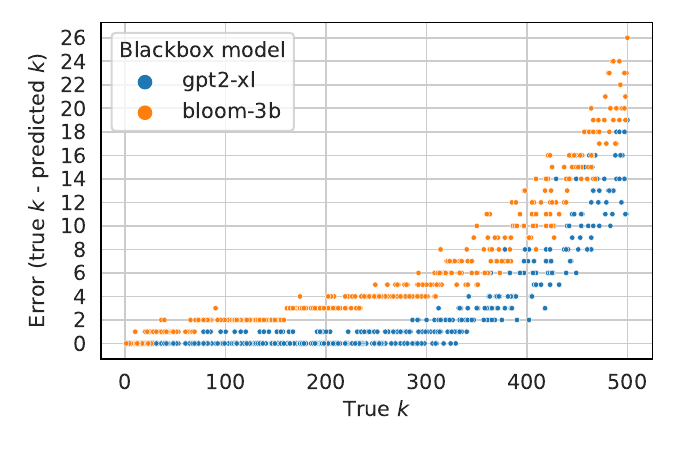}
    \caption{Error in top-$k$ estimation for 500 GPT-2 XL- and 500 bloom-3b-based systems. Error for both models is very low in common settings ($k$ < 100).}
    \label{fig:topk_error}
\end{figure}

\begin{table}[t]
    \centering
    \caption{Performance at $k$ estimation over 100 systems with $k$ values randomly chosen between 1 to 500. }
    \small
    \begin{tabular}{c|rrr}
        \toprule
        Model & Acc & Acc$\pm5$ & Avg Error \\
        \midrule
        GPT-2 Base & 28\% & 76\% & 1.3 \\
        GPT-2 XL & 44\% & 80\% & 0.9 \\
        BLOOM-3B & 0\% & 71\% & 2.3 \\
        pythia-2.7b & 22\% & 81\% & 1.1 \\
        \bottomrule
    \end{tabular}
    \label{tab:topk_estimation}
\end{table}

\subsection{Predicting $k$}
We used two prompt templates for estimating  top-$k$: \textsc{Nouns} and \textsc{Adverbs} (see Table \ref{tab:prompts}), each with 16 randomly selected exemplars.
We build an evaluation set of 100 systems, each with a $k$ selected uniform randomly to be between 1 and 500.
Table \ref{tab:topk_estimation} shows the accuracy of our approach on this evaluation set.\footnote{GPT-3 is omitted because the API does not expose top-$k$.}
We see that while our method is not so great at guessing $k$ perfectly, on average its guesses are between 0.9 and 2.3 off (depending on the underlying model).

In Figure \ref{fig:topk_error}, we plot accuracy as a function of true $k$ for GPT-2 XL.
This plot reveals that our method is especially effective at predicting $k$ for $k<300$, and accuracy deteriorates for higher $k$.
The vast majority of applications use $k$ well within this range, and it is simple to adjust for larger $k$ by increasing the max number of iterations parameter.

\begin{figure}[t]
    \centering
    \includegraphics[width=0.9\linewidth]{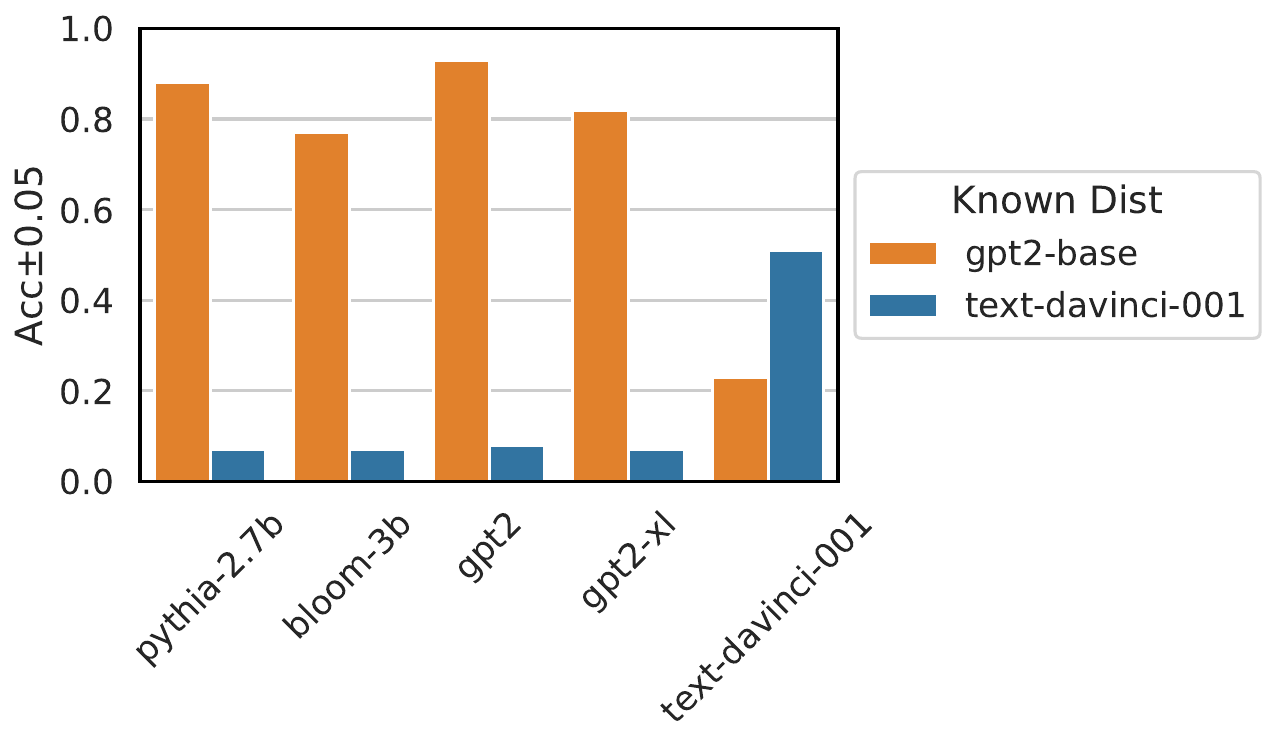}
    \caption{Performance at $p$ estimation over 100 systems with $p$ values ranging from 0.0 to 1.0, when the known distributions are computed using GPT-2 base (orange), and GPT-3 Davinci likelihoods (blue). Using Davinci as the known model leads to a better attack on GPT-3 models, but a worse one on all other models. RMSE in Table \ref{tab:topp_estimation}.}
    \label{fig:topp_estimation}
\end{figure}

\begin{figure}[t]
    \centering
    \includegraphics[width=\linewidth]{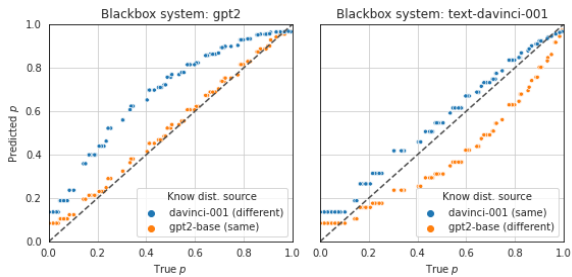}
    \caption{
    Mismatch between the known distributions and the distributions of the language model underlying the blackbox system lead to increased error. The matched estimates still have a slight systematic upward bias because we use the upper bounds for $p$ in our algorithm.
    }
    \label{fig:topp_scatterplot}
\end{figure}

\subsection{Predicting $p$}
We build an evaluation set of 100 systems, each with a randomly assigned $p$ between 0 and 1.
Table \ref{tab:prompts} shows the prompts used for top-$p$ estimation: \textsc{Months} and \textsc{Dates}.
For each prompt, we need to compute a known distribution over the next word.
We experiment with using both GPT-2 Base and GPT-3 Davinci for this.
For GPT-2, we compute the distribution directly; for GPT-3, we estimate it by running 1,000 trials with full random sampling.
Figure \ref{fig:topp_estimation} shows our method's accuracy at predicting within 0.05 of the true $p$ value.
Figure \ref{fig:topp_scatterplot} shows two limitations:
(1) our estimates are worse when there is significant mismatch between \gen's  distributions and our known distributions;
and (2) the minimum $p$ our method can predict is $\big(\text{Prob}(x_1^{(1)})+\text{Prob}(x_2^{(1)})\big)/2$, reducing accuracy for low $p$ values.
Further research is needed into the design of prompts which induce consistent distributions over many families of language models.

\begin{figure}
    \centering
    \includegraphics[width=0.8\linewidth]{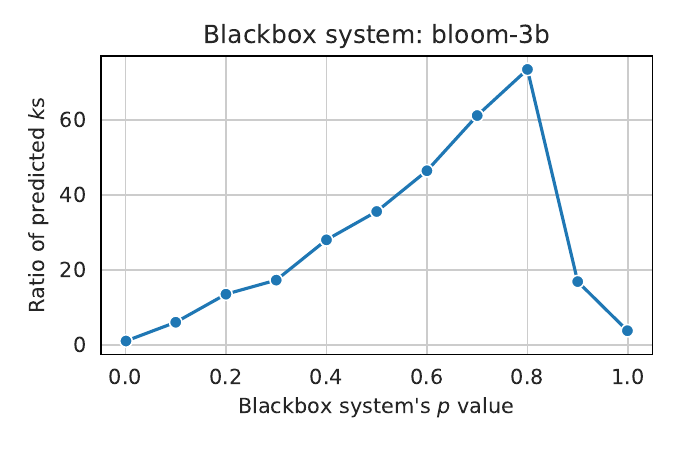}
    \caption{We plot the ratio between the $k$ estimated using \textsc{Adverbs} and using \textsc{Months}, for systems actually using top-$p$.
    Except when $p$ is near its extreme values, the prompt with larger $\Vp$ results in more generated words.}
    \label{fig:classification_experiment}
\end{figure}

\subsection{Distinguishing Top-$k$ and Top-$p$}

To evaluate our ability to distinguish between top-$k$ and top-$p$, we conduct the following experiment.
We take 10 systems with $p$ values ranging from 0.0 to 1.0 and find the chance we misclassify each system as using top-$k$.
Figure \ref{fig:classification_experiment} reports the results of this experiment.
We see that it would be fairly easy to distinguish the two methods by thresholding the ratio of the $k$ values returned by the two prompts.
Note that at the extreme values of $k$ and $p$, the method are indistinguishable.
Top-$k$ with $k$=1 and top-$p$ with $p$=0.0 are both the same as argmax; top-$k$ wirh $k$=$|\mathcal{V}|$ and top-$p$ with $p$=1.0 are both the same as full random sampling.

\subsection{ChatGPT}
We cannot repeat the exact same experiments with ChatGPT because (1) it does not use the exact prompt passed to the UI as the language model input, instead preprocessing it into a conversation format; and (2) the rate limiting prevents us from easily conducting many trials.
We instead employ four conversational-format prompts (see Table A\ref{tab:prompts_chatgpt}).
For the known distribution, we try out empirical distributions from five different versions of GPT-3, and take the one with lowest total variation distance from ChatGPT's output distribution.
Table \ref{tab:chatgpt} shows the $p$ estimated using each prompt.
Averaging these, we get $p$=0.81.
This estimate could be further narrowed down by incorporating more prompts, though of course we cannot validate this number due to opacity of the ChatGPT proprietary system.

\begin{table}[t]
    \centering
    \small
    \begin{tabular}{lr|r}
        \toprule
        Prompt & $n$ & Est. $p$ \\
        \midrule
        \textsc{MonthsChat} & 200 & 0.84 \\
        \textsc{DatesChat} & 125 & 0.74 \\
        \textsc{D20Chat} & 115 & 0.79 \\
        \textsc{D100Chat} & 500 & 0.86  \\
        \bottomrule
    \end{tabular}
    \caption{The values of $p$ estimated for ChatGPT using different prompts, where $n$ is the number of samples taken.}
    \label{tab:chatgpt}
\end{table}

\subsection{Room for Improvement}
All of the estimates reported in this section could be improved with additional queries to the model
For both $p$ and $k$ estimation, we average over the estimates from just two prompts, but using more prompts would lead to better estimates.
In addition, to improve top-$k$ estimation for larger $k$, one can increase the minimum number of times the least frequent items needs to be seen before the sampling loop terminates; in this paper, we set that value to 2.
Our methods could also be further improved by in-depth investigation of prompts which consistently produce close-to-uniform distributions across different families of language models.

Finally, while our methods do not currently address the case where temperature annealing is used in conjunction with top-$k$ or top-$p$, extending them to support this setting should be straightforward.
Temperature followed by top-$k$ is still top-$k$, and should be detectable via our methods.
Temperature followed by top-$p$ is trickier, because we no longer have a known distribution.
However, this combination can be detected by comparing the empirical distribution against a set of known distributions for common models; if the distribution does not match any of them, then we can conclude that either it is not using any known model, or that other distribution shaping such as temperature has been applied.

\section{Limitations}
Our method is limited to identifying when top-$p$ sampling or top-$k$ sampling is used.
We do not attempt to detect other decoding strategies which other systems might use. 
Additionally, there is no guarantee that a system would use a single decoding strategy---it is possible that different prompts may trigger different decoding strategies, or that A/B testing results in different users seeing different decoding strategies.
Our ChatGPT results were computed by two different authors on separate OpenAI accounts.
Also, we have no guarantees that the decoding strategy is not changed over time.
Some of our ChatGPT results were computed using the December 15, 2022 release while others were computed using the January 9, 2023 release.

Additionally, the biases in distributions that we see here could have other underlying reasons; for example, changes in the data can result in very different true distributions.

Furthermore, under the hood, an API might not be generating a new random generation each time an identical prompt is passed in.
Either random seeds might be getting re-used, or generations could be retrieved from a cache.
In both cases, the generations might look like argmax sampling.
It's also conceivable that certain combinations of fixed models could look like top-$k$/$p$.
For example, if a query is randomly routed to one of a series of $s$ servers, each serving a different model, we might interpret the decoding strategy to be top-$k$ even if each server is using argmax.
In these cases, an approach more like that proposed by \citet{tay2020reverse}, where classification of decoding strategy is made based on a long generated sequence (rather than single token system predictions, as in our approach), might be more effective.

For top-$p$ (though not top-$k$), we require access to an underlying distribution that approximates the model used.
This is not an issue for open source models or models with API access that allows specifying the decoding strategy, but it does limit the applicability of our method to newer proprietary models.
It may be possible to empirically determine distributions for carefully engineered prompts, but future work is needed for reverse engineering fully closed models.

\section{Conclusion}

Our attack shows that with even a little work, it is possible to reverse-engineer common decoding strategies.
Although we have focused here only on top-$p$ and top-$k$ sampling, these approaches generalize readily to other common methodologies when the output probability distributions are well-approximated.
Along with other recent work on reverse-engineering other parts of a language generation system \cite{zhang2023prompts}, it seems is infeasible to hide inference implementation details given black-box access to the system.

\bibliography{anthology,custom}

\clearpage
\appendix
\label{sec:appendix}
\onecolumn

\section{Algorithm for Estimating $k$}
\begin{algorithm}
\small
\caption{Algorithm for estimating $k$.}
\label{alg:topk}
\begin{algorithmic}
\State Given a system $\gen: m \mapsto r$ that takes an input prompt $m$ and outputs a single response token $r$, let $m_1$ and $m_2$ be two prompts that with high probability return responses from large ($\gg k$) sets of \textit{different} sizes---e.g. $m_1$ returns random nouns and $m_2$ returns random adverbs. 
\Function{\textsc{estimateK}}{Prompts $m_1, m_2; \gen : m \mapsto r$}
    \State $ \text{samples1} \mapsfrom [], \text{samples2} \mapsfrom []$
    \While{max number iterations not reached} \hspace{10em} // we set max iterations=32
        \For{$\{1..100\}$} \hspace{19.7em} // at most 3200 samples generated
            \State $\text{samples1}\text{.insert}(\gen(m_1))$
            \State $\text{samples2}\text{.insert}(\gen(m_2))$
        \EndFor
        
        \State $k_1 \mapsfrom \text{\# unique items in samples1}$
        \State $k_2 \mapsfrom \text{\# unique items in samples2}$
        
        \State $\text{minSamples} \mapsfrom (\text{samples1}^{(k_1)}>1$ and
        \State \hspace{2em} $\text{samples2}^{(k_2)}>1)$ \hspace{16em} // Boolean testing all items appear twice
        \If{$k_1=k_2$ and minSamples}
            \State \textbf{break}
        \EndIf
    \EndWhile
    \State \Return $\lfloor(k_1 + k_2) / 2\rfloor$ \hspace{18em} // guesses average if convergence not reached
\EndFunction
\end{algorithmic}
\end{algorithm}

\label{app:topk}
As we mentioned in the main paper text, suppose, for a given prompt $\prompt$, we call $\gen(\prompt)$, $n$ number of times, each time keeping just the first token of the output.
We can trivially lower bound $k$ by observing the number of unique items in a set of responses.
As $n$ approaches $\infty$, all $k$ allowed responses will be observed.
Since this is infeasible, the adversary's goal is to estimate $k$ while keeping $n$ as small as possible.

It is easy to see that the ideal prompt $\prompt$ is one that gives responses that are perfectly uniform over the entire vocabulary $\mathcal{V}$.
In the uniform case, we are left with the standard coupon collector problem~\cite{polya1930wahrscheinlichkeitsaufgabe}.
We would recover the exact value of $k$ with probability at least $1- \frac{1}{k}$ by setting $n>2 k\log k$.
Unfortunately, such a prompt is exceedingly difficult to engineer (see Appendix \ref{sec:promptselection}).

It turns out we can do almost as well without needing full uniformity.
The key building block for our attack is the construction of an $\prompt$ that distributes substantial probability mass onto a subset $\Vp \subseteq \mathcal{V}$ of the token space.
We require that for any $ k < |\Vp|$, we have $\text{Prob} \left( \gen(\prompt)=x^{(k)}  \right) \ge \frac{1}{ck} $, for some small constant $c$.
Put in plain language, we want to ensure that for any number of tokens $k$ the distribution might be truncated at, the least likely token that can be generated is no more than $c$ times less likely to appear than if the distribution were truly uniform.
If $n \ge 2ck \log (ck)$, then with probability at least $1-\frac{1}{ck}$, our prediction is exactly correct.
This result is far from tight, but follows easily from coupon collector on a uniformly random set of size $ck$.

In practice, we use Algorithm \ref{alg:topk}, which repeatedly estimates a lower bound for $k$ using two different prompts $\prompt_1$ and $\prompt_2$ for increasing numbers of trials until (1) the two estimates match and (2) the $x^{(k)}$ token appears at least twice in both generations (to prevent spurious matching).
In such a case, the expected number of trials $n$ is approximately bounded above by $2ck \log (ck)$ via coupon collector\footnote{Aside: the constant 2 that appears in the expected number of trials is due to requiring that the $k$th most likely token appears at least twice. However, it is unrelated to the constant $2$ that appears in the bound in the previous paragraph, which is chosen to ensure the $\frac{1}{ck}$ failure probability.}.

\section{Algorithm for Estimating $p$}
\begin{algorithm}
\caption{Algorithm for estimating $p$.}
\small
\label{alg:topp}
\begin{algorithmic}
\State Consider a language model $f_\theta: (m, r) \to \mathbb{R}$ that scores a prompt/response pair and a system $\gen: m \mapsto r$ that takes an input prompt $m$ and outputs a single response token $r$ using $f_\theta$ and top-$p$ sampling. Let $m_1$ and $m_2$ be two prompts that return responses from known distributions over relatively small sets ($|\Vp|$around 10-40)---e.g. $m_1$ returns random months and $m_2$ returns random dates within the month of March.
\Function{\textsc{estimateP}}{Prompts $m_1, m_2; \gen : m \mapsto r, f_\theta$}
\State $p_1 \mapsfrom \textsc{helper}(m_1, \gen, f_\theta)$
\State $p_2 \mapsfrom \textsc{helper}(m_2, \gen, f_\theta)$
\State \Return $(p_1 + p_2) / 2$
\EndFunction

\Function{\textsc{helper}}{Prompt $m; \gen, \text{known LM } f_\theta$
}
    \State $\text{baseProbs} \mapsfrom []$ \hspace{20.1em} // Will store known probability distribution
    \For{$v \in \mathcal{V}_m$} \hspace{20.3em} // $\Vp$ is the subset of tokens we consider
        \State $\text{baseProbs.insert}(P_{f_\theta}(r=v| m))$ \hspace{11.5em} // Probabilities using full random sampling
    \EndFor
    \State Sort baseProbs from largest to smallest.
    \State baseProbs.insert($\sum_{v \in \mathcal{V} - \mathcal{V}_m}\text{Prob}_{f_\theta}(r=v| m)$ \hspace{7.5em} // Summed probabilities of all out-of-set tokens
    
    \State $ \text{samples} \mapsfrom []$
    \For{$\{1..N\}$}
        \State $\text{samples}\text{.insert}(\gen(m))$
    \EndFor
    \State $l \mapsfrom \text{num unique items in samples}$
    
    \State $p \mapsfrom \sum_{i=1}^{l} \text{baseProbs}[i]$ 
    \State \Return $p$
\EndFunction

\end{algorithmic}
\end{algorithm}
Our goal is to estimate $p$ to within a factor of $\epsilon$.
This would be trivial to do if we could construct a prompt $m$ that is uniform over a subset $\Vp \subseteq \mathcal{V}$ of size at least $\frac{1}{\epsilon}$.
Then estimating $p$ would be equivalent to estimating top-$k$ for $k \approx \frac{p}{\epsilon}$ because each unique token seen implies a probability mass of $\epsilon$.

It is impossible to design a prompt which yields a perfectly uniform distribution.
However, although uniformity is desirable, for top-$p$ estimation, it is more important that the distribution of $\gen(m)$ is known, i.e., we have access to the underlying language model $f_\theta$.
If $k$ distinct tokens appear in the $p$-truncated distribution, then (using the same notation as above), we can bound $p$ as:
\[
\sum_{l=1}^{k-1} \text{Prob}_{f_\theta}(x^{(l)}) < p \le \sum_{l=1}^k \text{Prob}_{f_\theta}(x^{(l)}).
\]
Thus, given a known distribution, the top-$p$ reverse engineering problem reduces to top-$k$.

Even if we do not know exactly the underlying model for a blackbox system, we can construct prompts that appear to often return distributions close to a family of known distributions.
Then the error in estimating $p$ is just determined by how far off our guess of distribution is from the true underlying one.
Note that to ensure robustness against an imperfectly guessed distribution, we estimate $p$ using the sum of the $k$ largest in-vocabulary probabilities, rather than trying to actually match the probabilities for the unique items sampled.
This turns out to be important when prompts including exemplars are used, as the exemplars often create a bias in the tokens returned, but the overall drop-off in probabilities of most to least likely tokens tends to be more consistent.
However, for distribution matching, we use the actual distributions over tokens.


In this paper, we set a goal of determining $p$ to within $\epsilon = 0.05$ and construct two prompts with almost-known distributions over $k = 13$ and $k = 31$.
Because our guessed distributions are not perfect, instead of relying on a single distribution to bound our estimate, we instead average over the two upper bounds of $p$ derived from the different prompts and return that as our guess.
Additionally, as $k$ is small for both prompts, instead of using the stopping criterion of Algorithm 1, for each prompt, we always generate 3000 samples.
This means that with very high likelihood, we correctly return all possible items from the prompt's vocabulary.
Algorithm \ref{alg:topp} gives our implementation.

\newpage 
\section{Prompt Selection}
\label{app:prompt_engineering}

This Appendix gives more details on selecting good prompts for the decoding strategy detection task.

\section{Challenges}
\label{sec:promptselection}
We encountered many challenges in selecting appropriate prompts.
Our initial aim was to find prompts that induced an as-close-to-uniform distribution over the next token as possible.
In addition to the prompts decided on for our main experiments (Table \ref{tab:prompts}), we tried prompts meant to elicit a uniform distribution over digits, letters, dice rolls, and alphanumeric characters.
For some of these prompt styles, the main difficulty was in getting the language model to assign higher probability to the expected outputs for the prompt than to unexpected outputs.
For example, a prompt designed to elicit random digits would result in ``and'' being a more likely next token than several of the digits.
For other prompts, the distribution was not as random as we would have expected.
If exemplars were involved, even if they were chosen completely randomly, the model would try to follow any patterns observed in the exemplars.
For example, if a prompt containing randomly selected exemplars of digits happened to end with ``2 4 6'', then ``8'' would be by far the most likely next token.
Our difficulty here conforms with prior work that has shown that language models have significant biases toward certain numbers and words, even in settings where there should not be such bias.

\begin{table}[h]
    \centering
    \small
    \caption{Prompts showcasing the sensitivity of models to different exemplar choices. The exemplars, shown in blue, can be varied in order and count.}
    \label{tab:prompts_v1v2}
    \begin{tabular}{p{0.41in}|p{1in}| c}
        \toprule
        Name & Prompt v2 &  |$\mathcal{V}_p$| \\
        \midrule
        \textsc{Digits}  & Digits: \textcolor{blue}{4, 3, 2} & 10\\
        \midrule
        \textsc{ABC}  & Letters: \textcolor{blue}{E, F, P} & 26\\
        \bottomrule
    \end{tabular}
\end{table}

\begin{figure}
    \centering
    \includegraphics[width=\textwidth]{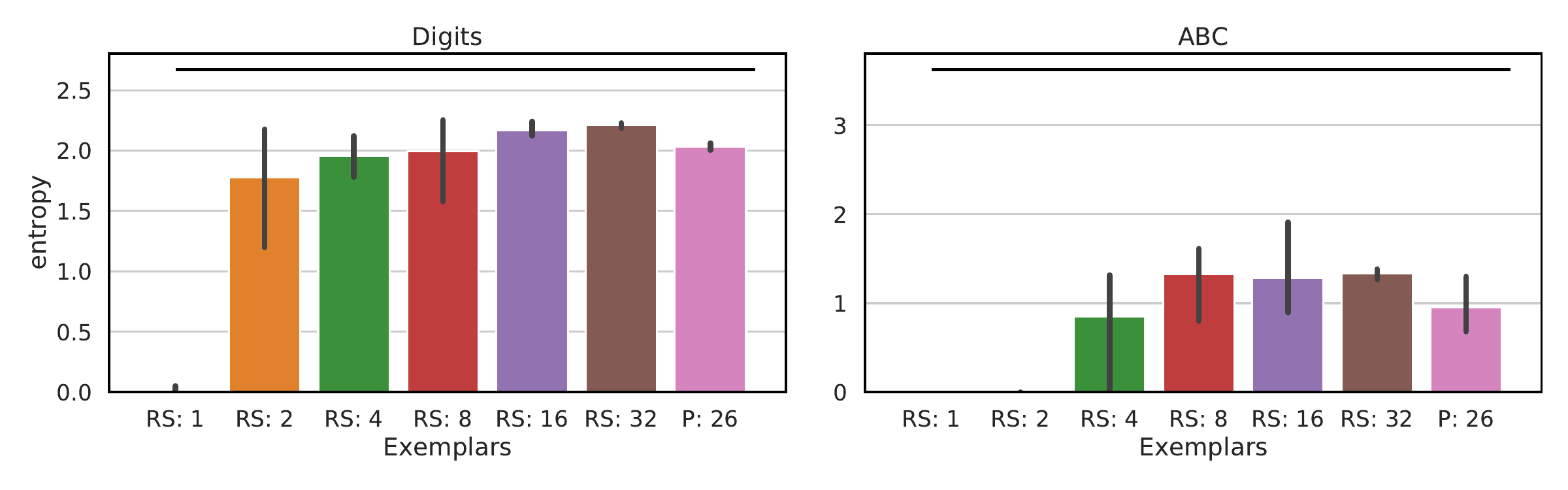}
    \caption{
    For each prompt style (\textsc{Digits} or \textsc{ABC}), we prompted with either [1, 2, 4, 8, 16, 32] exemplars selected randomly (RS), or with a random permutation of all expected outcomes (P).
    We ran three trials for each number of exemplars and generated the next word 5,000 times per trial. 
    The majority of next-word generated were within the vocabulary, however, sometimes they were not, in those cases, we discarded that output. 
    Missing bars indicate that there weren't enough generated next-words that were in the vocabulary to compute entropy.
    }
    \label{fig:entropy_dist}
\end{figure}

Figure \ref{fig:entropy_dist} shows the variance in output distributions for two exemplar-based prompts, \textsc{Digits} and \text{ABC} (Table \ref{tab:prompts_v1v2}), across different numbers of exemplars and different random selection of exemplars.
The \textsc{Digits} prompt is expected to output digits [0-9] with equal likelihood, and the random letters prompt is expected to output the letters [A-Z] with equal likelihood.
While with enough exemplars, the \textsc{Digits} prompt yielded consistently high entropy (i.e., close to uniform-random) distributions, the \textsc{Abc} prompt did not consistently improve with more exemplars.
In the end, we decided to avoid these prompts, and others which were too dependent on choice of exemplars.

\begin{figure*}[h]
    \centering
    \begin{subfigure}{0.5789\textwidth}
      \centering
      \includegraphics[width=\linewidth]{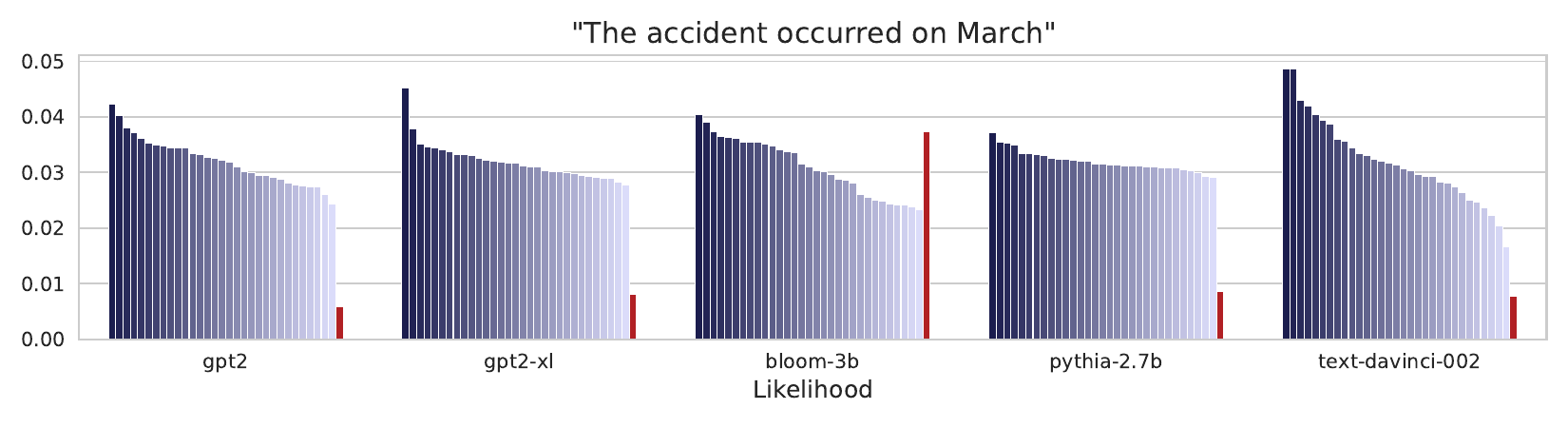}
    \end{subfigure}%
    \begin{subfigure}{0.42105\textwidth}
      \centering
      \includegraphics[width=\linewidth]{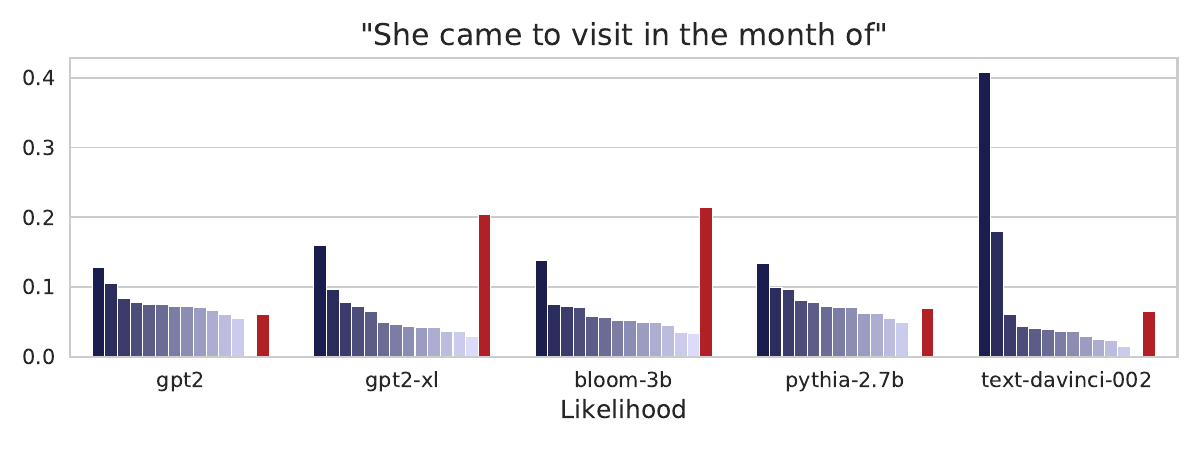}
    \end{subfigure}
    \caption{
    The likelihoods for the digits \{1-31\} given the prompt ``The accident occurred on March" (left) and for \{January thru December + Ramadan\} given the prompt ``She arrived in the month of'' (right), ordered from most to least likely. The sum probability of the remaining items in $\mathcal{V}$ is shown in red.
    }
    \label{fig:distributions}
\end{figure*}

For prompts to be used in top-$p$ estimation, one additional challenge is that ideally the prompt should yield a similar distribution when inputted to all popular language models.
As discussed in the paper, our estimates for $p$ are worse when there is a greater mismatch between the known distribution used for top-$p$ estimation and the true distribution of the language model underlying the blackbox system being attacked.
Figure \ref{fig:distributions} shows the known distributions for the two prompts we used in top-$p$ estimation, across several different models.
We see that some models have much spikier distributions than others.
The best approach (and the one we used to attack ChatGPT) is to choose the known distribution to use for top-$p$ estimation by keeping around a database of distributions from a bunch of different models, and then comparing the output distribution from the blackbox system to each distribution in the database.
We can then choose to estimate $p$ using the known distribution with the lowest relative entropy with the blackbox's one.


\subsection{Chosen Prompt Details}
Here we describe the actual prompts used in our experiments.
For the \textsc{Nouns} and \textsc{Adverbs} prompts, we assumed access to the GPT-2 vocabulary and used Spacy (\texttt{en\_core\_web\_sm}) to identify all tokens in the vocabulary corresponding to nouns and adverbs.
In all experiments with these prompts, we used 16 randomly selected exemplars from these lists.
An example prompt for \textsc{Nouns} is: ``List of nouns chosen completely randomly: negativity diarrhea problems eloqu money aspect vertex fraternity stone breast skies pushes probabilities ink north creditor''.
In our experiments estimating top-$k$, for each system being evaluated, we varied the random seed, resulting in a slightly different prompt.
We did this to avoid any systematic biases resulting from always using the same choice of exemplars.
For the non-exemplar-based prompts, we did not assume vocabulary access and instead relied on the expectation that letters, digits, and common words are present in most model vocabularies.

As mentioned in the main paper, different prompts were needed to attack ChatGPT than for the experiments on open-source models because ChatGPT expects its inputs to be in a conversation format and it does not offer control over the number of words generated (without careful prompt design, it tends to return tens to hundreds of words).
Table \ref{tab:prompts_chatgpt} gives the prompts used to attack ChatGPT.

\begin{table}[]
    \centering
    \small
    \caption{Prompts uses to estimate the $p$ value for ChatGPT.}
    \label{tab:prompts_chatgpt}
    \begin{tabular}{p{0.8in}|p{2.2in}|p{0.1in}}
        \toprule
        Name & Prompt & |$\Vp$| \\
        \toprule
        \textsc{MonthsChat} & write one word for the rest of this sentence: ``She came to visit in the month of'' & 13 \\
        \midrule
        \textsc{DatesChat} & write one word for the rest of this sentence: ``The accident occured on March'' & 31 \\
        \midrule
        \textsc{D20Chat}  & write one number for the rest of this sentence: ``I rolled a D20 and the outcome was'' & 20 \\
        \midrule
        \textsc{D100Chat} & Could you roll me a D100? We're playing D\&D. Answer with just the roll value and nothing else. & 100 \\
        \bottomrule
    \end{tabular}
\end{table}

\section{Scientific Artifacts}
We use the following language models in our research:
\begin{itemize}
    \item \textbf{BLOOM 3B}: This model was released by BigScience under the RAIL License v1.0 with the goal to ``to enable public research on large language models'' \citep{scao2022bloom}. It can be downloaded at \url{https://huggingface.co/bigscience/bloom}.
    \item \textbf{Pythia 2.7B}: This model was released by EleutherAI under the MIT License with the goal of enabling research on ``interpretability analysis and scaling laws'' \citep{2023pythia}. It can be downloaded at \url{https://github.com/EleutherAI/pythia}.
    \item \textbf{GPT-2 base and XL}: These models were released by OpenAI under the MIT license with the goal of fostering language model research \citep{radford2019language}. They can be downloaded at \url{https://huggingface.co/gpt2}.
    \item \textbf{ChatGPT and GPT-3 model family}: These models are only available via OpenAI's API or through OpenAI's web interface. Our experiments with them fall under OpenAI's research policy, found at \url{https://openai.com/api/policies/sharing-publication/#research-policy}.
\end{itemize}

We chose these models evaluate on because (1) we wanted to evaluate our method on a wide range of independently trained models using different paradigms and training dataset choices.
For example, though we conduct all our experiments using English prompts, we can observe the impact of BLOOM being trained on multilingual data, in that for the \textsc{Months} prompt, BLOOM puts significant probability-mass on non-English month names, which could affect our $p$ estimates for BLOOM models. 

\section{Computational Resources}
Preliminary experiments were run in Google Colab using a Pro membership, which gave access to one Tesla T4. Subsequent experiments were running on a Google Cloud machine with 8 Tesla V100s.
No more than 100 hours were spent running computation on this machine, which has a cost of \$17 per hour.

\section{Additional Results}
Table \ref{tab:topp_estimation} gives the numbers used in Figure \ref{fig:topp_estimation} in the main paper, as well as the root mean-square error between the true and estimated $p$ values.

\begin{table}[h]
    \centering
    \small
    \caption{Performance at $p$ estimation across 100 estimations with $p$ values random from 0 to 1.
    On the left, GPT-2 Base was used to compute known distributions, and on the right GPT-3 was used to compute the known distributions.
    }
    \begin{tabular}{c|rr|rr}
        \toprule
        & \multicolumn{2}{|c|}{GPT-2 Base} & \multicolumn{2}{|c}{GPT-3 Davinci v1} \\
        Model & Acc$\pm .05$ & RMSE & Acc$\pm .05$ & RMSE \\
        \midrule
        GPT-2 Base & 0.93 & 0.03 & 0.08 & 0.19 \\
        GPT-2 XL & 0.82 & 0.04 & 0.07 & 0.21 \\
        Davinci v1 & 0.23 & 0.14 & 0.51 & 0.06  \\
        BLOOM-3B & 0.77 & 0.04 & 0.07 & 0.22 \\
        pythia-2.7b & 0.88  & 0.03 & 0.07 & 0.20 \\
        \bottomrule
    \end{tabular}
    \label{tab:topp_estimation}
\end{table}

\end{document}